\documentclass[preprint]{imsart}

\usepackage{fullpage}
\usepackage{amsmath}
\usepackage{amssymb}
\usepackage{amsthm}
\usepackage{graphicx}
\usepackage{color}
\usepackage{subfig}
\usepackage{macros}
\usepackage{microtype}
\usepackage{url}

\usepackage[top=1.5in, bottom=1.5in, left=2in,
  right=2in]{geometry}

\newcommand{\MethodName}{SinkProp}

\urldef\homepage\url{http://www.cs.toronto.edu/~rpa}

\begin{document}
\begin{frontmatter}
  \title{Ranking via Sinkhorn Propagation}

  \author{\fnms{Ryan Prescott} \snm{Adams}\thanksref{t1}%
    \ead[label=e1]{rpa@cs.toronto.edu}}%
  \and
  \author{\fnms{Richard S.} \snm{Zemel}%
    \ead[label=e2]{zemel@cs.toronto.edu}}%
  \address{Department of Computer Science\\
    University of Toronto}%
  \address{\printead{e1,e2}}

  \thankstext{t1}{\homepage}
  
  \runauthor{R.P. Adams and R.S. Zemel }
  \runtitle{ Ranking via Sinkhorn Propagation}
  
\begin{abstract}
  It is of increasing importance to develop learning methods for
  ranking.  In contrast to many learning objectives, however, the
  ranking problem presents difficulties due to the fact that the space
  of permutations is not smooth.  In this paper, we examine the class
  of \emph{rank-linear} objective functions, which includes popular
  metrics such as precision and discounted cumulative gain.  In
  particular, we observe that expectations of these gains are
  completely characterized by the marginals of the corresponding
  distribution over permutation matrices.  Thus, the expectations of
  rank-linear objectives can always be described through locations in
  the Birkhoff polytope, i.e., doubly-stochastic matrices (DSMs).  We
  propose a technique for learning DSM-based ranking functions using
  an iterative projection operator known as Sinkhorn normalization.
  Gradients of this operator can be computed via backpropagation,
  resulting in an algorithm we call Sinkhorn propagation, or
  \emph{\MethodName{}}.  This approach can be combined with a wide range of
  gradient-based approaches to rank learning.  We demonstrate the
  utility of \MethodName{} on several information retrieval data sets.
\end{abstract}

\end{frontmatter}

\section{Introduction}
\label{sec:intro}
The task of ranking is straightforward to state: given a query and a
set of documents, produce a ``good'' ordering over the documents based
on features of the documents and of the query.  From the point of view
of supervised machine learning, we define the ``goodness'' of a
ranking in terms of graded relevances of the the documents to the
query, using an objective function that rewards orderings that rank
the more relevant documents highly.  The task is to take a training
set of queries in which the documents are labeled with known
relevances, and build an algorithm that is capable of producing
orderings over queries in which the relevance labels are unknown.

There are several aspects of this problem, however, that make it
difficult to train a ranking algorithm and which have recently led to
a flurry of insightful, creative approaches.  The first thorny aspect
of the learning-to-rank problem is that, unlike typical supervised
classification and regression problems, we wish to learn a
\emph{family} of functions with varying domain and range.  This
difficulty arises from the property that queries may have varying
numbers of documents; the set of input features scales with the number
of documents and the size of the output ordering also changes.

The second difficulty in learning to rank is that the space of
permutations grows rapidly as a function of the number of documents.
A fully Bayesian decision-theoretic approach to the ranking problem
would construct a rich joint distribution over the relevances of
documents in a query and then select the optimal ordering, taking this
uncertainty into account.  While this is feasible for small queries,
this optimization quickly becomes intractable.  It is therefore more
desirable to develop algorithms can directly produce permutations,
rather than relevances.

Finally, any objective defined in terms of orderings over
relevance-labeled documents is necessarily piecewise-constant with
respect to the parameters of the underlying function.  That is, it may
not be possible to compute the gradient of a training gain in terms of
the parameters that need to be learned.  Without such a gradient, many
of the powerful machine learning function approximation tools, such as
neural networks, become infeasible to train.

In this paper, we develop an end-to-end framework for supervised
gradient-based learning of ranking functions that overcomes each of
these difficulties.  We examine the use of doubly-stochastic matrices
as differentiable relaxations of permutation matrices, and define
popular ranking objectives in those terms.  We show that the expected
value of an important class of ranking gains are completely preserved
under the doubly-stochastic interpretation.  We further demonstrate
that it possible to propagate gradient information backwards through
an \emph{incomplete Sinkhorn normalization} to allow learning of
doubly-stochastic matrices.  We also show how this leads to
flexibility in the choice of pre-normalization matrices and enables
our approach to be integrated with other powerful ranking methods,
even as the number of per-query documents vary.  We note in particular
that this approach is well-suited to take advantage of the recent
developments in the training of deep neural networks.

\section{Optimizing Expected Ranking Objectives via Doubly-Stochastic Matrices}
\label{sec:dsm}

In a learning-to-rank problem, the training data are~$N$ sets, called
\textit{queries}, the~$n$th of which has size~$J_n$.  The items within
each of these query sets are called \textit{documents}.  The features
of document~$j$ in query~$n$ are denoted as~${\bx^{(n)}_j\in\mcX}$.
In the training set, each document also has a \textit{relevance}
label~${r^{(n)}_j\in\{0,1,2,\ldots,R\}}$, where~$R$ indicates the
maximum relevance.  We denote the vector of relevances for query~$n$
as~$\br^{(n)}$.

A ranking of the documents for a given query can be represented as a
permutation, mapping each document $j$ to a rank $k$.  Each
permutation is an element of $\mcS_J$, the symmetric group of
degree~$J$.  The aim then is to learn a family of functions that
output permutations: ${f_J:\mcX^J\to\mcS_J}$,
for~${J\in\{1,2,\ldots\}}$, where, as above,~$\mcX$ is a set of
features associated with each of the~$J$ items.

\subsection{Ranking Objective Functions}

The most critical component when learning to rank is the definition of
an objective function which identifies ``good'' and ``bad'' orderings
of documents.  For a query of size~$J$ with known relevances, we
identify three well-studied scoring functions over the group~$\mcS_J$:
the order-$K$ \textit{normalized discounted cumulative gain}
(NDCG@$K$) \cite{jarvelin-kekalainen-2002a}, the order-$K$
\textit{precision} (P@$K$) \cite{MAP}, and the
\textit{rank biased precision} (RBP) \cite{moffat-zobel-2008a}.

Writing~$s[k]$ to indicate the index of the document at rank~$k$ in
permutation~$s$, NDCG@$K$ is given by
\begin{align*}
  \mcL_{\textsf{NDCG@}K}(s\,;\,\br) &=
  \frac{\displaystyle\textrm{DCG@}K(s\,;\,\br)}
  {\displaystyle\max_{s'}\textrm{DCG@}K(s'\,;\,\br)}
  \\
  \textrm{DCG@}K(s\,;\,\br) &= \sum_{k=1}^K
  \mcG(r_{s[k]})\,\mcD(k)\\
  \mcG(r) &= 2^r-1
  \\
  \mcD(k) &= \begin{cases} 1 & \text{if~${k=1,2}$}\\
    \log_2(-k) &\text{if ${k > 2}$}
    \end{cases}.
\end{align*}
The P@$K$ objective, for binary relevances is
\begin{align*}
\mcL_{\textsf{P@}K}(s\,;\,\br) 
&= \frac{1}{K}\sum^K_{k=1}r_{s[k]}
\end{align*}
and RBP is
\begin{align*}
\mcL_{\textsf{RBP}}(s\,;\,\br) &=
(1-\alpha)\sum^J_{k=1}r_{s[k]} \alpha^{k-1},
\end{align*}
where~${\alpha\in[0,1]}$ is a ``persistence'' parameter.  The natural
training objective, then, is to find a family of functions~${\mcF =
  \{f_J : J\in\{1, 2, \ldots\}\}}$ that maximizes the aggregate
empirical gain, subject to some regularization penalty~$Q(\mcF)$:
\begin{align}
\label{eqn:empirical-loss}
\mcF^\star &= \argmax{\mcF=\{f_J\}}\left\{
-Q(\mcF) +
\sum^N_{n=1} \mcL(f_{J_n}(\{\bx^{(n)}_j\}^{J_n}_{j=1})\,;\,\{r^{(n)}_j\}^{J_n}_{j=1})
\right\}.
\end{align}

\subsection{Doubly-Stochastic Matrices as Marginals Over Permutations}

As discussed in Section~\ref{sec:intro}, one of the difficulties with
the objective in \eqnref{eqn:empirical-loss} is that it is
discontinuous.  That is, changes in~$f_J$ only effect the training
data via the discrete ordering, and so are piecewise-constant modulo
the regularization~$Q(\mcF)$.  One way to address this is to replace
the objectives of the previous section with expectations of these
objectives, where the expectations are with respect to a distribution
over rankings~\cite{softrank}.  That is, given a distribution
over permutations, denoted by~$\psi(s)$, and the relevances~$\br$, the
expected gain is
\begin{align}
  \label{eqn:psi-loss}
  \mathbb{E}_{\psi}[\mcL(\br)]
  &= \sum_{s\in\mcS_J}\psi(s)\,\mcL(\br).
\end{align}
Many ranking objectives, however, are characterized by element-wise
sums over the associated permutation matrix~$\bS$,
i.e.,~${S_{j,k}\!=\!\delta_{j,s[k]}}$, where~$\delta_{j,k}$ is the
Kronecker delta function.
We call such objectives \emph{rank-linear}, and they have the general form
\begin{align*}
  \label{eqn:rank-linear}
  \mcL(s\,;\,\br) &= \sum^J_{j=1}\sum^J_{k=1} S_{j,k}\,\ell(r_j,k).
\end{align*}
This has been observed previously in the literature, e.g.,
\cite{le-smola-2007a,chapelle-wu-2010a}.  One remarkable aspect of
rank-linear objectives, however, is that the expectation in
Eq.~(\ref{eqn:psi-loss}) is completely captured by the marginal
probability that~${S_{j,k}\!=\!1}$:
\begin{align}
  \mathbb{E}_{\psi}[\mcL(\br)] &=
  \sum^J_{j=1}\sum^J_{k=1}
  \ell(r_j,k)\,\mathbb{E}_{\psi}[S_{j,k}].
\end{align}
The entry-wise marginal distributions necessary to describe this
expectation form a doubly-stochastic matrix (DSM).  A DSM is a
nonnegative square matrix in which each column and each row sum to
one.  Permutation matrices are special cases of doubly-stochastic
matrices.  By Birkhoff's Theorem, every doubly stochastic matrix can
be expressed as the convex combination of at most~${J^2\!-\!2J\!+\!2}$
permutations (see, e.g.,~\cite{bapat-1997}).  It is natural, then, to
think of a DSM as a relaxation of a permutation that incorporates
precisely the uncertainty that is appropriate for rank-linear gains.
That is, the~$j$th row of a DSM provides a properly-normalized
marginal distribution over the rank of item~$j$.  As the columns must
also sum to one, this set of~$J$ distributions is consistent in the
sense that they cannot, e.g., attempt to place all documents in the
first position. While the DSM does not indicate which permutations
have non-zero probability, it does provide all the information that is
necessary to determine the expected gain under rank-linear objectives.

We leverage this interpretation of DSMs as providing a set of
consistent marginal rank distributions in the expected ranking
objective.  Let ${\bPi\in[0,1]^{J\times J}}$ be a doubly stochastic
matrix.  We interpret entry~$\Pi_{j,k}$ as the marginal probability
that item~$j$ is at rank~$k$.  We use this to define a differentiable
objective function
\begin{align}
  \mathbb{E}_{\bPi}[\mcL_{\textsf{NDCG@}K}(\br)] &=
  \frac{\sum^J_{j=1} \sum^K_{k=1} \Pi_{j,k}\,
    \mcG(r_j)\,\mcD(k)}
  {\max_{s'}\mcL_{\textsf{DCG@}K}(s'\,;\,\br)}
  \label{eqn:surr-ndcg}
\end{align}
that is the expectation of NDCG@$K$ under an (unknown) distribution
over permutations that has the marginals described by~$\bPi$. 
The expected P@$K$ can be defined similarly:
\begin{align*}
  \mathbb{E}_{\bPi}[\mcL_{\textsf{P@}K}(\br)] &=
  \frac{1}{K}\sum^J_{j=1}r_j\sum^K_{k=1}\Pi_{j,k},
\end{align*}
as can the expectation of rank biased precision:
\begin{align*}
\mathbb{E}_{\bPi}[\mcL_{\textsf{RBP}}(\br)] &=
(1-\alpha)
\sum^J_{j=1}r_j\sum^J_{k=1}
\Pi_{j,k} \alpha^{k-1}.
\end{align*}
Note that all three of these expectations recover their original
counterparts when~$\bPi$ is a permutation matrix, which could then be
thought of as providing a set of degenerate marginals.

\subsection{Choosing a Single Permutation}
\label{sec:choose}
Although we have defined differentiable rank-linear objectives in
terms of doubly-stochastic matrices, our test-time task remains the
same: we must produce a single ordering.  Under our ``consistent
marginals'' interpretation of doubly-stochastic matrices, the natural
objective is the one the maximizes the log likelihood:
\begin{align}
  \label{eqn:cost}
  s^\star &= \argmax{s\in\mcS_J}
  \sum^J_{k=1} \ln\Pi_{s[k],k}.
\end{align}
This maximization is a bipartite matching problem and can be solved
in~$\bigO(J^3)$ time using the Hungarian algorithm
\cite{munkres-1957a}.  For queries of more than a few hundred
documents, this cubic time complexity becomes a bottleneck.  To speed
up selection of a permutation, we use a ``short-cut'' bipartite
matching scheme that uses a quadratic algorithm to compute a global
ordering and then uses the Hungarian algorithm on only the
top~${P\!<\!J}$ documents under the global ordering, resulting
in~$\bigO(K^2\!+\!P^3)$ running time.

In the first pass, we compute the expected rank of each document under
the marginal distributions implied by the doubly-stochastic matrix:
\begin{align*}
  \mathbb{E}_{\bPi}[\text{rank}_j] &= \sum^K_{k'=1}\Pi_{j,k'}k'.
\end{align*}
These expected ranks can be sorted to compute an ordering~$\hat{s}$
for all~$J$ documents.  In the second phase, the Hungarian algorithm
is applied to the top~$P$ documents and the
submatrix~$\bPi_{\hat{s}[1:P],1:P}$ using the score
in~Eq.~(\ref{eqn:cost}).  This provides an improved ordering for the
top~$P$ documents, while keeping the remainder of the permutation
fixed.

While this procedure offers no theoretical guarantees for the
optimality of the global matching, it is well-suited to the ranking
problem.  First, metrics such as NDCG are most heavily influenced by
the top-ranked documents in the ordering; focusing the expensive
computations on this subset is a sensible heuristic.  Second, the main
way that this procedure would act pathologically is if the row-wise
distributions were highly multimodal, i.e., significant mass split
between very high and very low ranks.  For the types of functions we
explore in Section~\ref{sec:params}, however, this kind of behavior
is unlikely.

\begin{figure}[t]
  \centering%
  \subfloat[Initial matrix]{%
    \centering%
    \includegraphics[width=0.23\textwidth]%
    {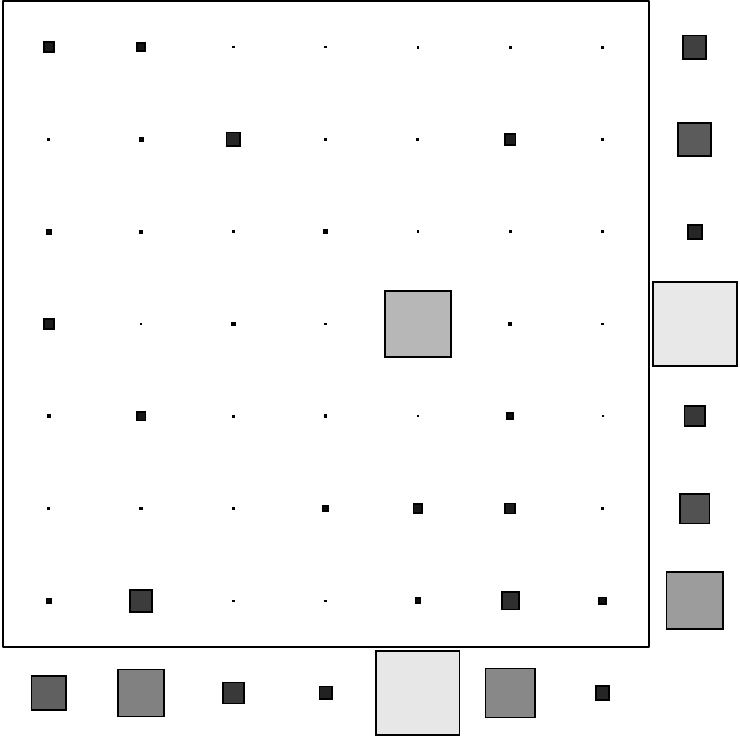}%
  }~\quad%
  \subfloat[First iteration]{%
    \centering%
    \includegraphics[width=0.23\textwidth]%
    {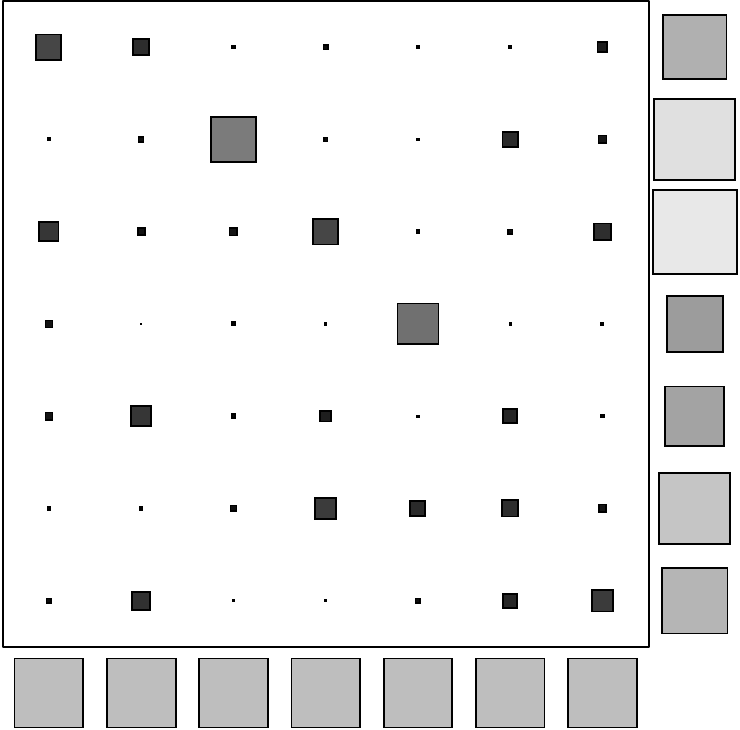}%
  }~\quad%
  \subfloat[Second iteration]{%
    \centering%
    \includegraphics[width=0.23\textwidth]%
    {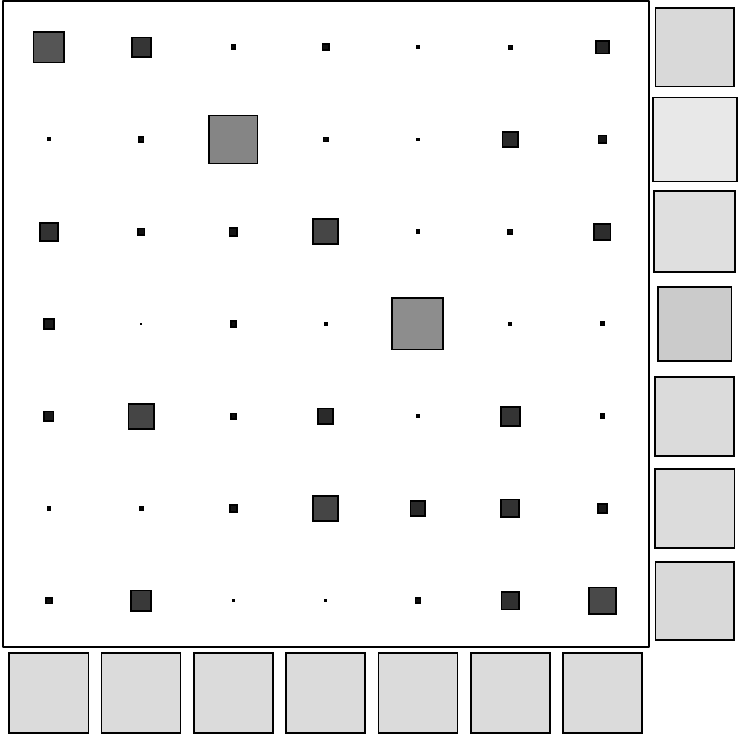}%
  }~\quad%
  \subfloat[Third iteration]{%
    \centering%
    \includegraphics[width=0.23\textwidth]%
    {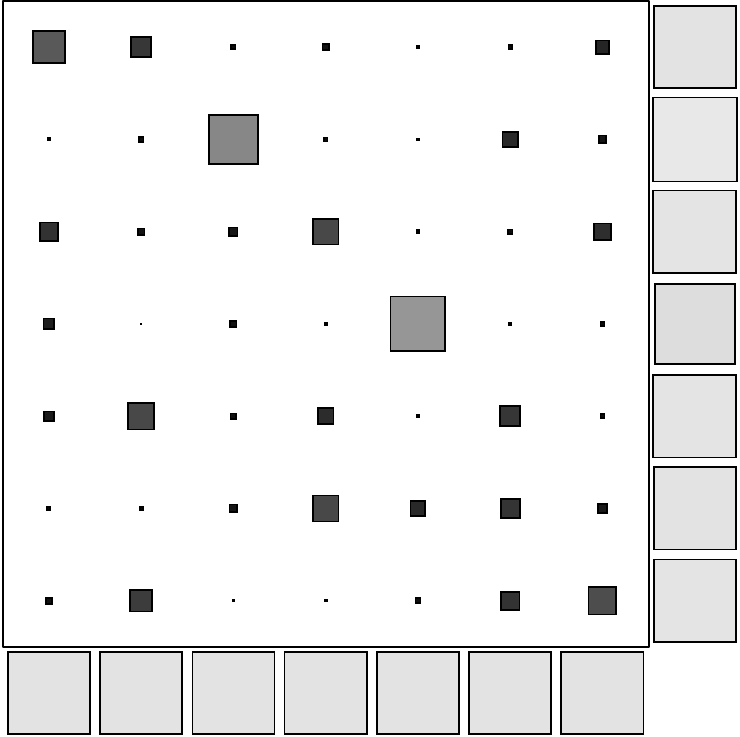}%
  }
  \caption{Hinton diagrams of three iterations of Sinkhorn
    normalization.  The row and column sums are shown as squares on
    the right and bottom of the box.  The matrix quickly balances:
    in~(d) the row- and column-wise sums are almost
    indistinguishable.}
  \label{fig:normalize}
\end{figure}

\section{Learning to Rank with Sinkhorn Gradients}
\label{sec:sinkhorn}
Having relaxed our objective functions from permutations to
doubly-stochastic matrices, we are no longer seeking to learn
functions~${f_J:\mcX^J\to\mcS_J}$ (which output permutations), but
instead a family of functions~${g_J:\mcX^J\to\mcW_J}$, where~$\mcW_J$
is the set of~${J\times J}$ doubly stochastic matrices (the Birkhoff
polytope).  Such functions are difficult to construct, however, as it
is not straightforward to parameterize doubly-stochastic matrices.
This is in contrast to, for example, right-stochastic matrices for
which we can easily construct a surjective function using a row-wise
softmax.  There is, however, an iterative projection procedure known
as a \textit{Sinkhorn normalization} \cite{sinkhorn}, which takes a
nonnegative square matrix and converts it to a doubly-stochastic
matrix by repeatedly normalizing rows and columns.  More formally, we
define row and column normalization functions:
\begin{align*}
  \mcT_{\textsf{R}}(\bA) &= \bA \; \oslash (\bA \bone \bone^\trans) &
  \mcT_{\textsf{C}}(\bA) &= \bA \; \oslash ( \bone \bone^\trans \bA),
\end{align*}
where~$\oslash$ is the Hadamard (elementwise) division and~$\bone$ is a
vector of ones.  We can then define the iterative function
\begin{align*}
\mcZ^{i}(\bA) &=
\begin{cases}
\bA & \text{if $i = 0$}\\
\mcT_{\textsf{R}}(\mcT_{\textsf{C}}(\mcZ^{i-1}(\bA))) & \text{otherwise}
\end{cases}.
\end{align*}
The function~${\mcZ^{\infty}:\reals_{+}^{J \times J}\to\mcW_J}$ is a
Sinkhorn normalization operator and, when it converges, it produces a
doubly-stochastic matrix.  

Sinkhorn and Knopp \cite{sinkhorn-knopp-1967} presented necessary and
sufficient conditions for convergence of this procedure to a unique
limit.  In summary, convergence is guaranteed for most non-negative
matrices, but some non-negative matrices cannot be converted into
doubly stochastic matrices because of their pattern of zeros
(expressed as conditions on zero minors in the original
matrix). $\bigO(V |\log \epsilon|)$ Sinkhorn steps suffice to reach
$\epsilon$-near double stochasticity if all the matrix entries are
in~${[1, V]}$~\cite{sinkhorn-convergence}. Hence most applications of
Sinkhorn normalization smooth the original matrix to facilitate
convergence.

In this paper, we introduce the idea of an \textit{incomplete}
Sinkhorn normalization, which are the functions~$\mcZ^{i}(\bA)$, for
${i < \infty}$.  The incomplete Sinkhorn normalization allows us to
define the objective functions of the previous section in terms of
square matrices whose only constraints are that the entries must be
nonnegative.  That is, if we can now produce a matrix~$\reals_{+}^{J
  \times J}$ for each training query, this can be approximately
normalized and the ranking functions can be evaluated.  Most
interestingly, however, it is possible to compute the gradient of the
training objective with regard to the initial unnormalized matrix.
This can be done efficiently by propagating the gradient backward
through the sequence of row and column normalizations, as in
discriminative neural network training.  We refer to this
backpropgation as \emph{\MethodName{}}.

As in backpropagation, in \MethodName{} we proceed through layers of
computation and we assume that the output of each layer provides the
input to a scalar function for which we are computing the gradient in
terms of the layer's inputs.  This function is here
denoted~$\mcU(\bA')$ and we assume that the
gradient~${\partial\mcU(\bA')/\partial\bA'}$ has already been
computed.  We can use this to compute the gradient of a row
normalization via
\begin{align*}
  \frac{\partial}{\partial A_{j,k}}\mcU(\mcT_{\textsf{R}}(\bA))
  &= \sum^J_{k'=1}  
  \left[
  \frac{\delta_{k,k'}}{\sum^J_{k''=1}A_{j,k''}}
  - \frac{A_{j,k'}}{\left(\sum^J_{k''=1}A_{j,k''}\right)^2}\right]
\frac{\partial\mcU(\bA')}{\partial A'_{j,k'}}.
\end{align*}
The gradient of the column normalization is essentially identical,
modulo appropriate transposition of indices.  The
function~$\mcU(\cdot)$ here corresponds to the composition of any
number of Sinkhorn normalizations with the rank-linear objective.
This enables us to use a small amount of code to backpropagate through
an arbitrary-depth incomplete Sinkhorn normalization.

\section{Parameterizing the Pre-Sinkhorn Matrix}
\label{sec:params}
We have so far defined a differentiable training function that allows
us to optimize a rank-linear objective in terms of an
unconstrained nonnegative square matrix.  The final piece of our
framework is to define a family of
functions~${h_J:\mcX^J\to\reals_{+}^{J\times J}}$,
for~${J\in\{1,2,\ldots\}}$.  There are several approaches that could
be taken to this problem.  We will focus on two examples in which the
functions~$h_J$ can be computed from~$J$ evaluations of a single
function~${\phi:\mcX\to\reals^D}$.

\subsection{Partitioned Probability Measures}
One approach to the functions~$h_J$ is to construct them in terms of
parametric probability densities~$\pi(u\given\theta)$ defined
on~$(0,1)$.  The parameters~$\theta$ are taken to be in~$\reals^D$ for
all~$J$, so that they may be computed via~$\phi(\bx)$.  For any~$J$,
we can define~$J$ equally-spaced bins in~$(0,1)$, with
edges~${\rho_J[j]=j/J}$, for~${j\in\{0,1,\ldots,J\}}$.  We then define
the ``output matrix'' from~$h_J$ to be
\begin{align}
  A_{j,k} &= \int^{\rho_J[k]}_{\rho_J[k-1]}\pi(u\given\theta_j =
  \phi(\bx_j))\,\mathrm{d}u.
\end{align}
In this construction, each row of the matrix arises by subdividing the
mass of the row-specific cumulative probability distribution function
whose parameterization is determined by~$\phi(\bx_j)$.  Intuitively,
this means that for any given row, the more mass that appears near
zero, the greater the preference for that document appearing higher in
the ranking.  Reasonable choices for~$\pi(\cdot)$ include the beta
distribution, the probit, and the logit-logistic (LL):
\begin{align*}
\pi_{\sf{LL}}(u\given\theta=\{\mu,\sigma\}) &=
\frac{1}{4\sigma}
\text{sech}^2\left\{\frac{\ln(u/(1-u))-\mu}{2\sigma}\right\}
\left(\frac{1}{u} + \frac{1}{1-u}\right).
\end{align*}
Both the PDF and CDF of the LL distribution are fast to compute.  This
overall construction is appealing as it is differentiable, and is a
natural approach to regression of variable-sized square matrices.

\subsection{Smoothed Indicator Functions}
\label{sec:indicator}
One deficiency of the approach of the previous section is that there
is no interaction between the documents except through the Sinkhorn
iterations.  An alternative approach, inspired by
SmoothRank~\cite{chapelle-wu-2010a}, is to construct the pre-Sinkhorn
matrix via:
\begin{align}
  A_{j,k} &= \exp\left\{
    -\frac{1}{2\sigma^2}
    \left(\phi(\bx_j) - \phi(\bx_{\hat{s}[k]})\right)^2
    \right\},
\end{align}
where~${D\!=\!1}$ and the permutation~$\hat{s}$ arises from sorting
the~$\phi(\bx_j)$.  In the limit of~${\sigma\to 0}$ this recovers the
permutation matrix implied by~$\hat{s}$.  In the
limit~${\sigma\to\infty}$, the matrix becomes all ones.  As discussed
in~\cite{chapelle-wu-2010a}, this function is continuous but not
differentiable.  However, it is almost everywhere differentiable, as
the discontinuities in the first derivatives only occur at ties
between the~$\phi(\bx_j)$.  This has no practical effect for
optimization purposes.

In both of these examples, it is the function~$\phi(\cdot)$ that is of
primary interest, and whose parameters are being optimized during
training.  Assuming that the document features are~$M$-dimensional
real-valued vectors, i.e., ~${\mcX=\reals^M}$, then a simple linear
function is a reasonable base case:~${\phi(\bx)=\bW \bx}$,
where~${\bW\in\reals^{M\times D}}$.  More interesting is the
possibility of using deep neural networks and other more sophisticated
function approximation tools.

\begin{figure}[t]
  \centering%
  \subfloat[HP2003]{%
    \includegraphics[width=0.24\textwidth]%
    {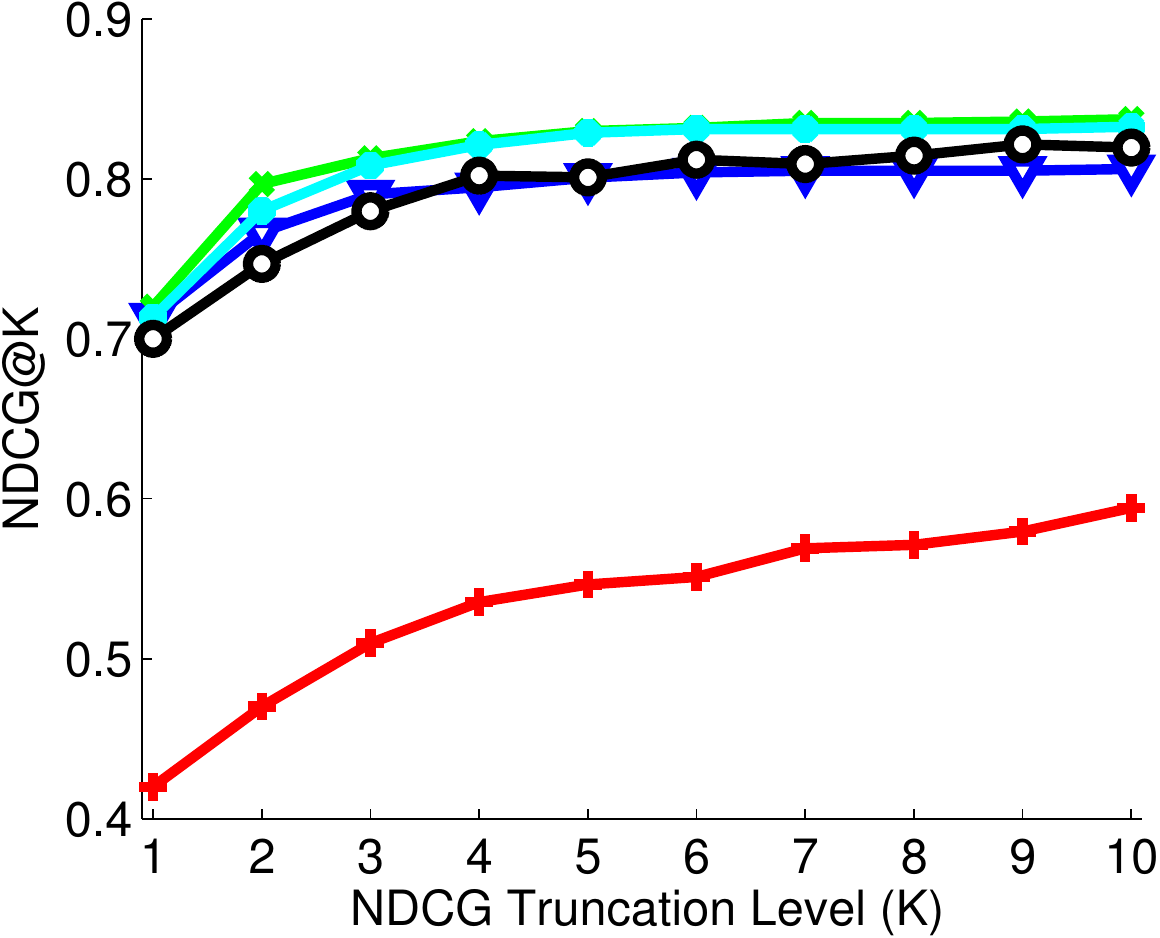}%
  }~%
  \subfloat[NP2003]{%
    \includegraphics[width=0.24\textwidth]%
    {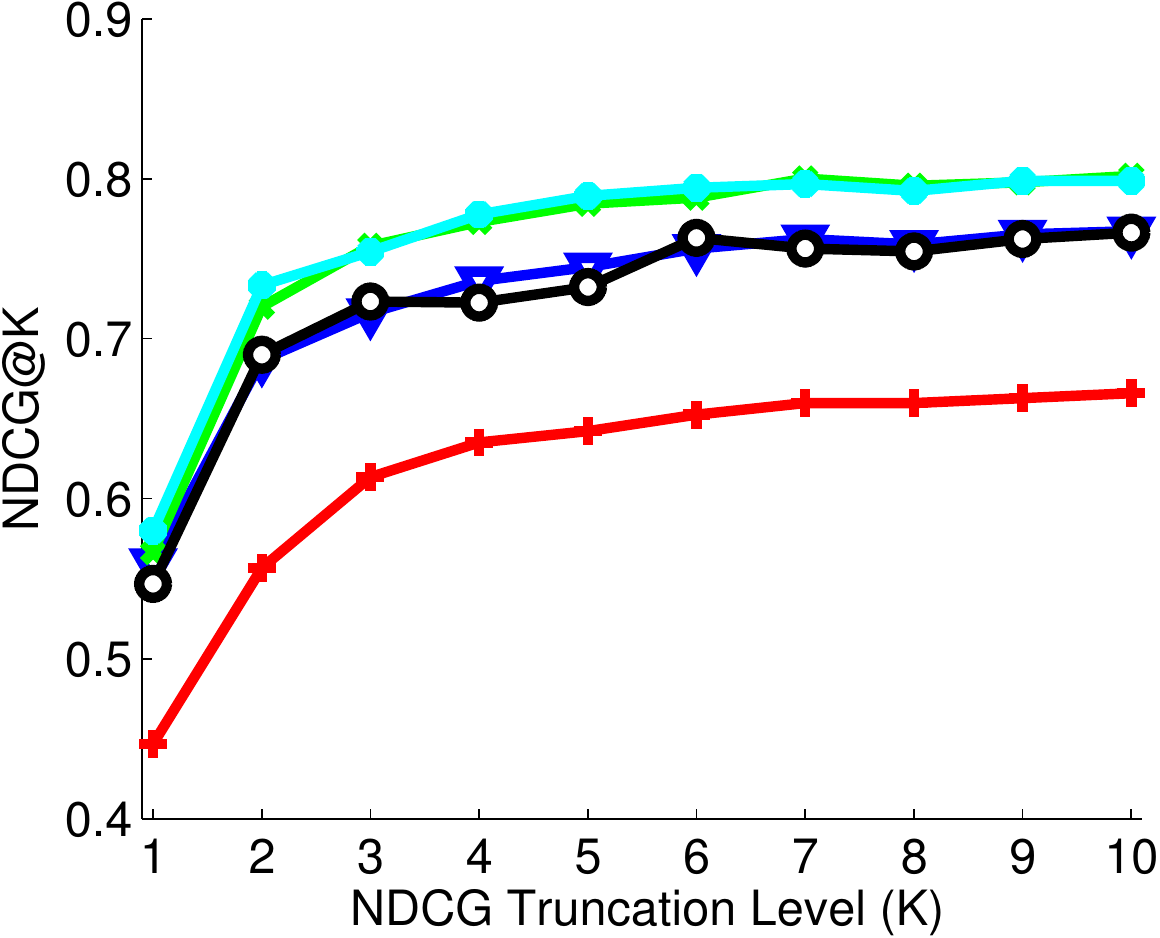}%
  }~%
  \subfloat[TD2003]{%
    \includegraphics[width=0.24\textwidth]%
    {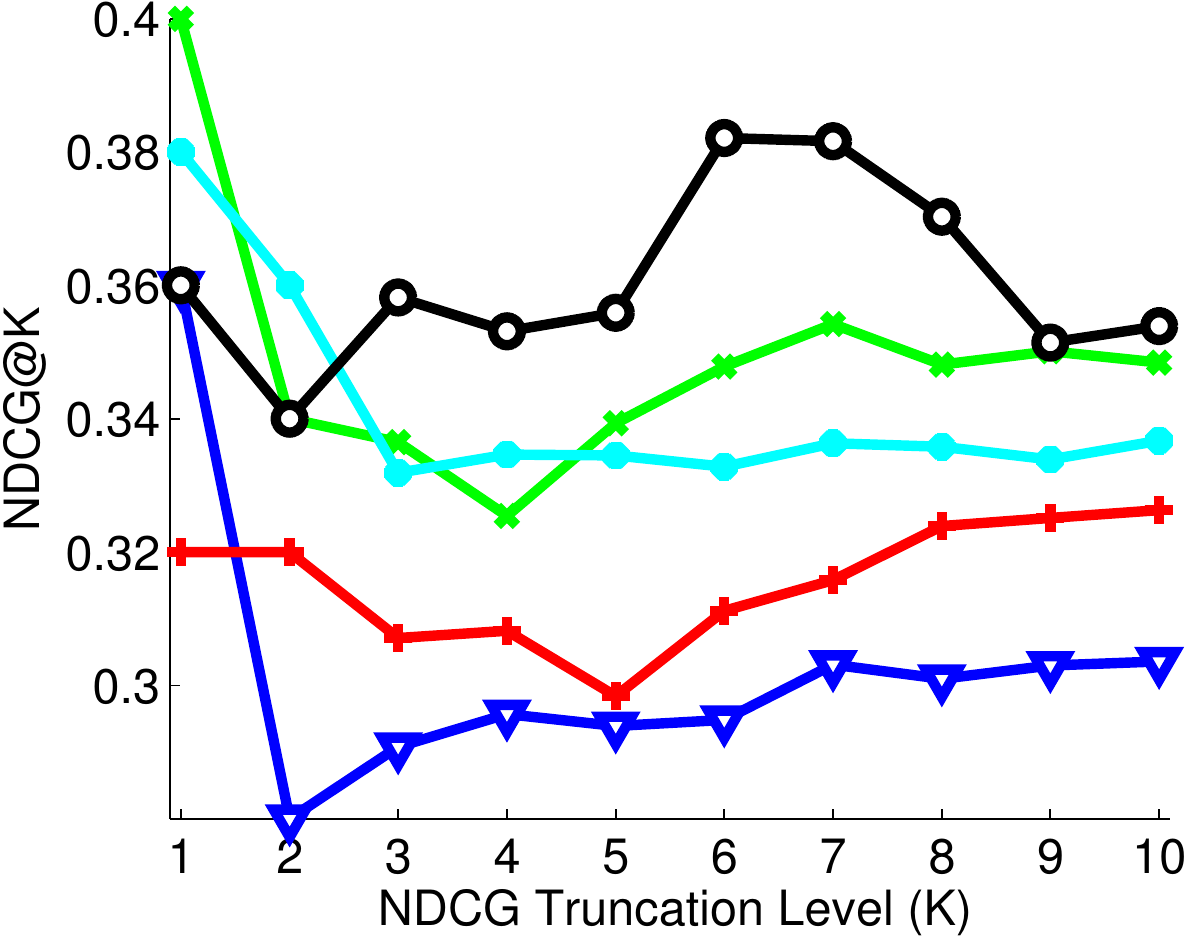}%
  }~%
  \subfloat[OHSUMED]{%
    \includegraphics[width=0.24\textwidth]%
    {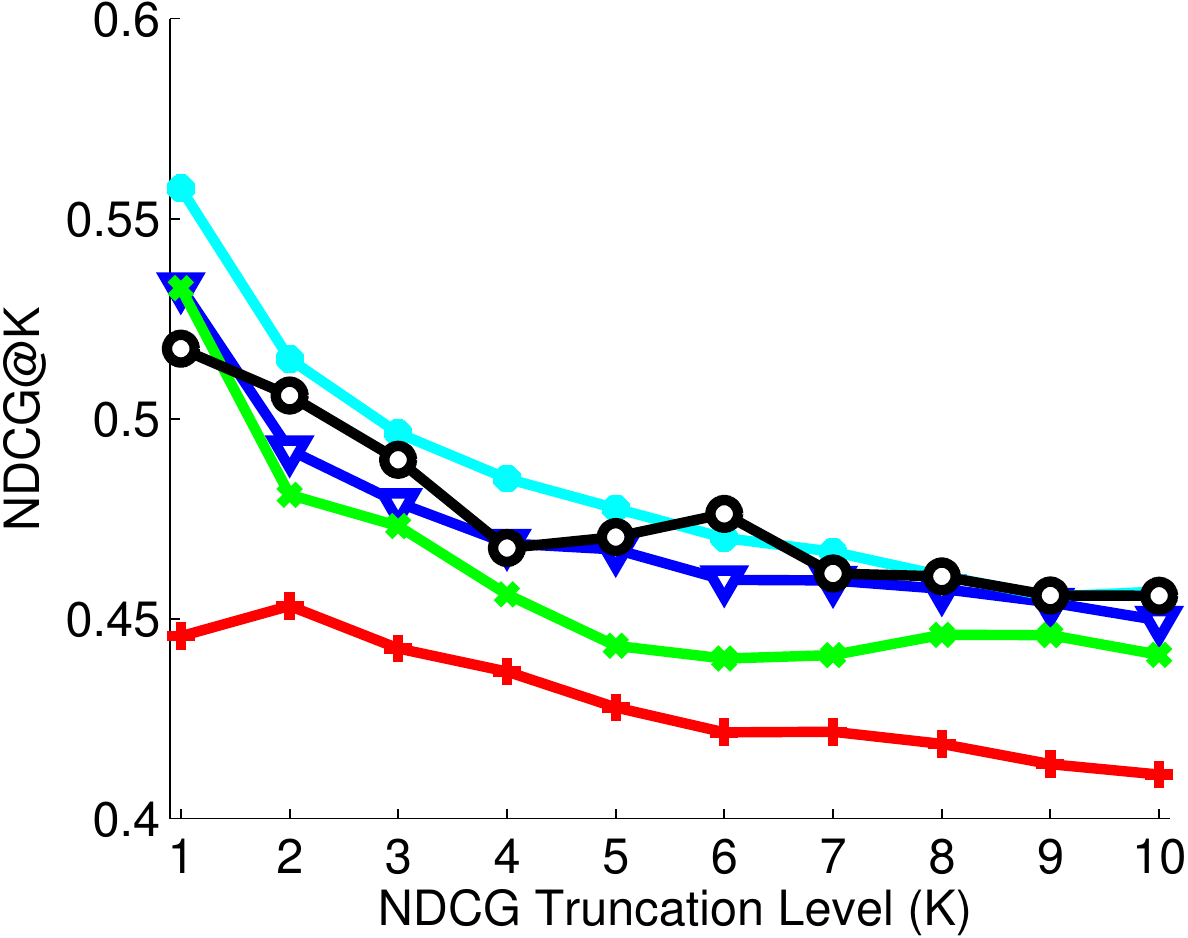}%
  }\\%
  \subfloat[HP2004]{%
    \includegraphics[width=0.24\textwidth]%
    {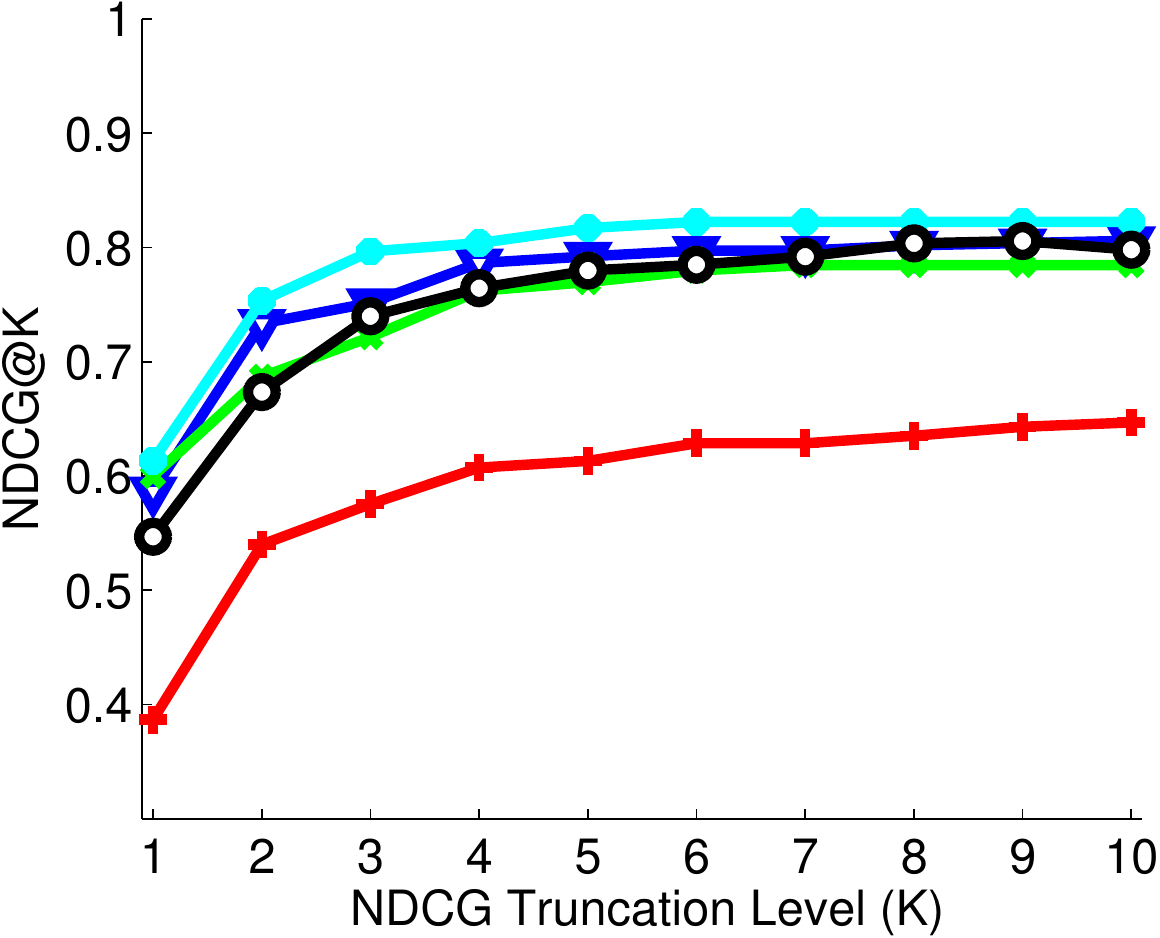}%
  }~%
  \subfloat[NP2004]{%
    \includegraphics[width=0.24\textwidth]%
    {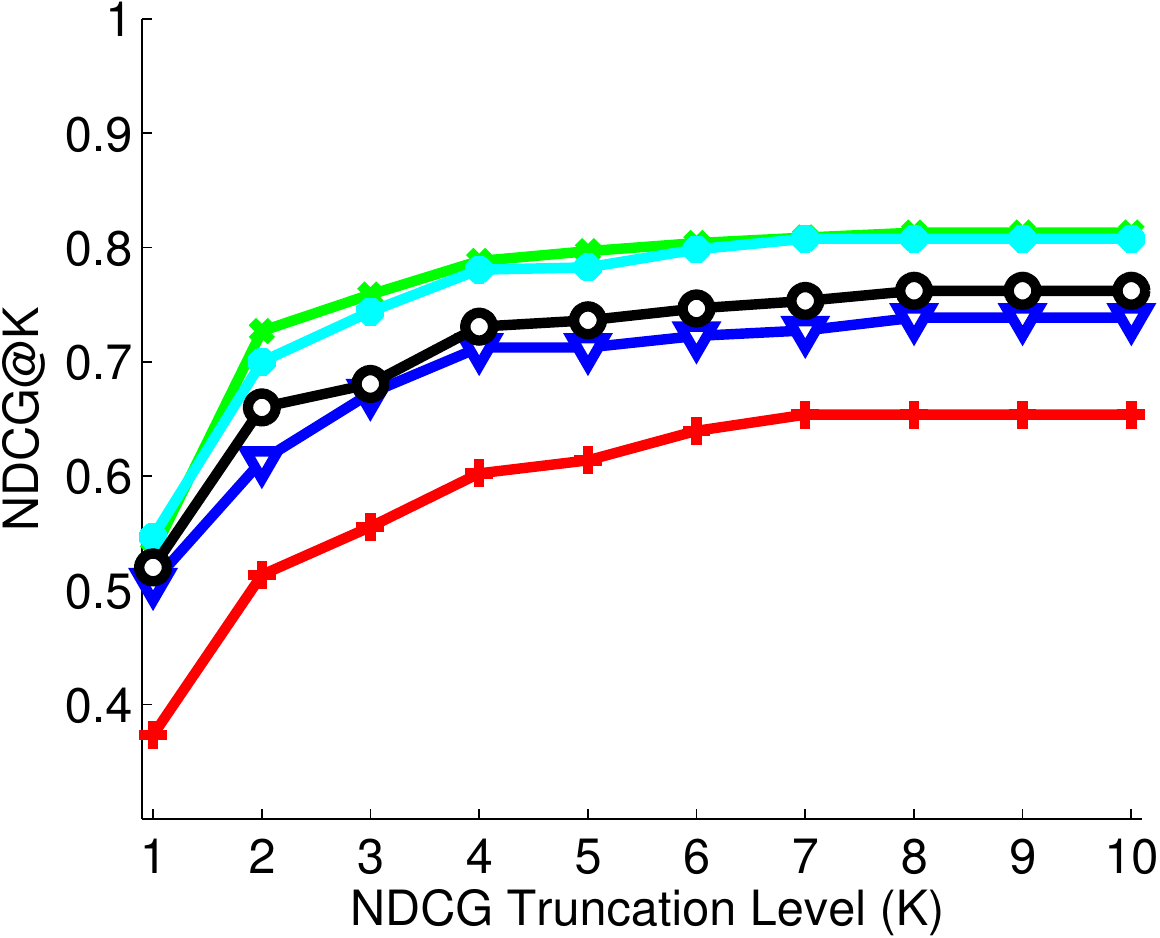}%
  }~%
  \subfloat[TD2004]{%
    \includegraphics[width=0.24\textwidth]%
    {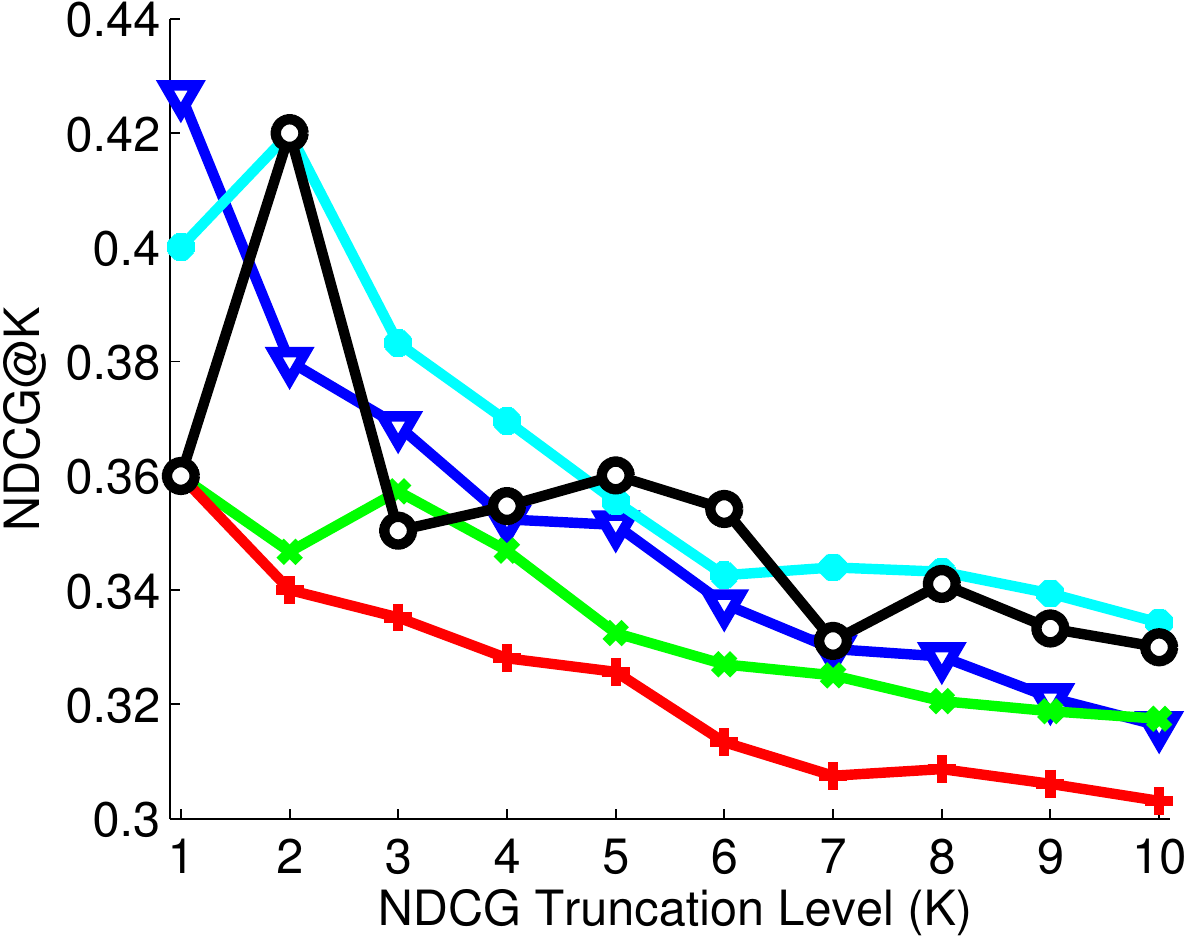}%
  }~%
  \subfloat{%
    \centering%
    \parbox[t]{0.24\textwidth}{%
      \qquad~%
      \includegraphics[width=0.17\textwidth]%
      {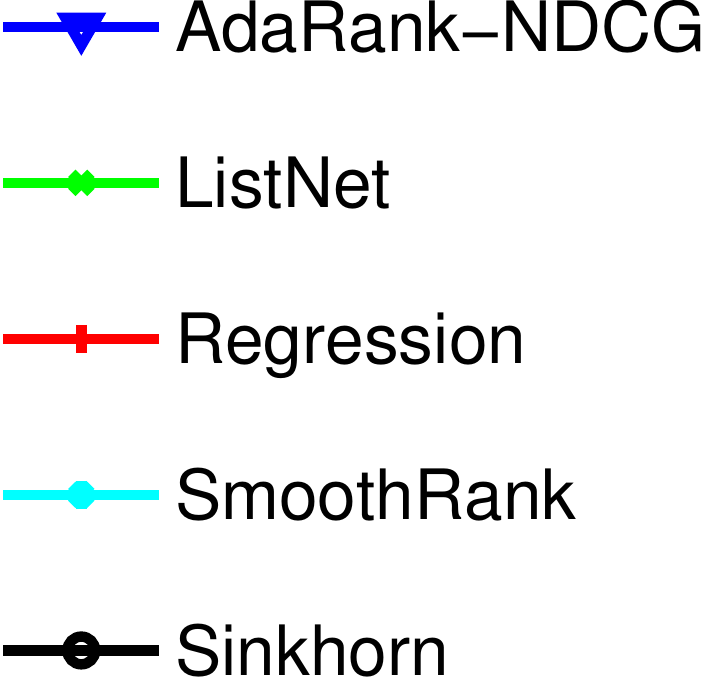}%
    }
  }
  \caption{Results on the seven LETOR 3.0 data sets.}
  \label{fig:letor-results}
\end{figure}

\section{Empirical Evaluation on LETOR Data}
\label{sec:letor}
In this section we report on comparisons with other approaches to
ranking, using seven data sets associated with LETOR 3.0 \cite{letor}
benchmark.  Following the standard LETOR procedures, we trained over
five folds, each with distinct training, validation and test splits.
We used the training objective of Eq.~(\ref{eqn:surr-ndcg}), with~$K$
set to be the number of documents in the largest query.  This achieved
the best performance in our experiments, similar to the results
reported in \cite{chapelle-wu-2010a}.  These experiments used the
smoothed indicator function approach described in
Section~\ref{sec:indicator}.  Optimization was performed using
L-BFGS\footnote{\url{http://www.cs.ubc.ca/~schmidtm/Software/minFunc.html}},
annealing the smoothing constant~$\sigma$ as in
\cite{chapelle-wu-2010a}.  We initialized with the MLE regression
weights under squared loss.  Early stopping was performed based on the
NDCG of predictions on the validation set.  We regularized the weights
using the~$\ell_2$ distance to the MLE regression weights, selecting
the regularization penalty using the validation set.  To reduce
training time, the training data were resampled into a larger number
of smaller queries.  Each initial query was turned into twenty derived
queries whose documents were sampled with replacement from the
original.  The number of documents in each derived query was Poisson
distributed, with mean determined by the original number of documents,
up to a maximum of 200.  Before performing the Sinkhorn normalization,
a small constant (${\approx 10^{-6}}$) was added to each entry in the
matrix.  Five Sinkhorn iterations were performed.  When specific rank
predictions were made at test time, the short-cut Hungarian method was
used as described in Section~\ref{sec:choose}, with~${P\!=\!200}$.

The results for the seven different data sets are shown in
Figure~\ref{fig:letor-results}.  Several publicly available
baselines\footnote{\url{http://research.microsoft.com/en-us/um/beijing/projects/letor/}}
are shown for comparison: AdaRank \cite{adarank}, ListNet
\cite{listnet}, SmoothRank \cite{chapelle-wu-2010a} and basic
regression.  In each figure, the testing NDCG score is shown as a
function of truncation level.  As can be seen in the graphs, the
Sinkhorn approach is generally competitive with the state-of-the-art.
On the TD2003, the Sinkhorn normalization appears to offer a
substantial advantage.

\section{Related Work}
\MethodName{} builds on a rapidly expanding set of approaches to rank
learning.  Early learning-to-rank methods employed surrogate gain
functions, as approximations to the target evaluation measure were
necessary due to its aforementioned non-differentiability.  More
recently, methods have been developed to optimize expectation of the
target evaluation measure, including
SoftRank~\cite{softrank}.
BoltzRank~\cite{BoltzRank} and
SmoothRank~\cite{chapelle-wu-2010a}.  These methods all attempt to
maximize the expected gain of the ranking under any of the gain
functions described above. The crucial component of each is the
estimate of the distribution over rankings: SoftRank uses a
rank-binomial approximation, which entails sampling and sorting ranks;
BoltzRank uses a fixed set of sampled ranks; and SmoothRank uses a
softmax on rankings based on a noisy model of scores.  \MethodName{}
can be viewed as another method to optimize expected ranking gain, but
the effect of the scaling is to concentrate the mass of the
distribution over ranks on a small set, which peaks on the single
chosen rank selected by the model at test time.

Sinkhorn scaling itself is a long-standing method with a wide variety
of applications, including discrete constraint satisfaction problems
such as Sudoku~\cite{sudoku-sinkhorn} and for updating probabilistic
belief matrices \cite{Balakrishnan04polynomialapproximation}. It has also been
used as a method for finding or approximating the matrix permanent, as the 
permanent of the scaled matrix is the permanent of original matrix
times the entries of the row and column scaling
vector~\cite{AcceptReject}.  Recently, Sinkhorn normalization has
been employed within a regret-minimization approach for on-line
learning of permutations~\cite{helmbold-warmuth-2009}.

Although this work represents the first approach to ranking that has
directly incorporated Sinkhorn normalization into the training
procedure, previously-developed methods have also found it useful.
The SoftRank algorithm \cite{softrank}, for example, uses Sinkhorn
balancing at test-time, as a post-processing step for the approximated
ranking matrix.  Unlike the approach proposed here, however, it does
not take this step into account when optimizing the objective
function.  SmoothRank uses half a step of Sinkhorn balancing,
normalizing only the scores within each column of the matrix, to
produce a distribution over items at a particular rank.

\section{Discussion and Future Work}
\label{sec:discussion}
In this paper we have presented a new way to optimize learning
algorithms under ranking objectives.  The conceptual centerpiece of
our approach is the observation that the expectations of certain kinds
of popular ranking objectives --- ones we have dubbed
\emph{rank-linear} --- can be evaluated exactly even if only marginal
distributions are available.  This means that the expected value of
these ranking objectives can be computed from the doubly-stochastic
matrix that arises from their location within the Birkhoff polytope.
To actually learn the appropriate doubly-stochastic matrices, it is
possible to apply a well-studied iterative projection operator known
as a Sinkhorn normalization.  Remarkably, gradient-based learning can
still be efficiently performed in the unnormalized space by
backpropagating through the iterative procedure, which we call
\emph{\MethodName{}}.  We demonstrated the effectiveness of this new
approach by applying it to seven information retrieval datasets.

There are several promising future directions for work in this area.
In the context of ranking, the ability to backpropagate gradients of
rank-linear objectives enables wide possibilities in retrieval of
non-text documents.  In particular, there have been significant recent
advances in gradient-based training of neural networks for
discrimination of images and speech (e.g.,
\cite{lee-etal-2009a,mohamed-dahl-hinton2009}).  The SinkProp approach
could enable these networks to be trained to produce permutations over
these more general types of objects.  More broadly, we have shown that
it is practical to backpropagate training gradients through iterative
projection operators.  Such operators can be defined for a variety of
structured outputs: matching problems and image correspondence tasks,
for example.  Finally, while the notion of rank-linearity is specific
to the problem of learning permutations, it seems likely that it can
be expanded to other types of structured-prediction tasks, leading to
efficient computation of expected gains under DSM-like
representations.

\subsection*{Acknowledgements}
The authors wish to thank Sreangsu Acharyya, Daniel Tarlow and Ilya
Sutskever for valuable discussions.  RPA is a junior fellow of the
Canadian Institute for Advanced Research.

\bibliographystyle{unsrt}
\bibliography{draft}

\end{document}